\documentclass[a4paper]{article}[11pt]
\usepackage{times,amsmath}
\usepackage{amssymb}
\usepackage{algorithm2e}
\usepackage{amsfonts}
\usepackage{graphicx}
\usepackage{graphics}
\usepackage{mathtools}
\usepackage{algpseudocode}
\usepackage{amsmath}
\usepackage{textcomp}
\usepackage{wrapfig,lipsum,booktabs}
\usepackage{multirow}
\usepackage[affil-it]{authblk}
\usepackage{fancyhdr}
\usepackage{epstopdf}
\usepackage{url}
\usepackage[named]{algo}

\begin{document}
\title{Use of Ensembles of Fourier Spectra in Capturing Recurrent Concepts in Data Streams}

\author{Sripirakas Sakthithasan\textsuperscript{*} , Russel Pears\textsuperscript{*}, Albert Bifet\textsuperscript{\#} and Bernhard Pfahringer\textsuperscript{\#}}
\affil{Auckland University of Technology, Auckland 1010, New Zealand\textsuperscript{*}}
\affil{Department of Computer Science, University of Waikato, New Zealand\textsuperscript{\#}}
\maketitle

\begin{abstract}
In this research, we apply ensembles of Fourier encoded spectra to capture and mine recurring concepts in a data stream environment. Previous research showed that compact versions of Decision Trees can be obtained by applying the Discrete Fourier Transform to accurately capture recurrent concepts in a data stream.  However, in highly volatile environments where new concepts emerge often, the approach of encoding each concept in a separate spectrum is no longer viable due to memory overload and thus in this research we present an ensemble approach that addresses this problem. Our empirical results  on real world data and synthetic data exhibiting varying degrees of recurrence reveal that the ensemble approach outperforms the single spectrum approach in terms of classification accuracy, memory and execution time.

\end{abstract}

\section{Introduction}
In many real world applications, patterns or concepts recur over time. Machine learning applications that model, capture  and recognize concept re-occurrence gain significant efficiency and accuracy advantages over systems that simply re-learn concepts each time they re-occur. When such applications include safety and time critical requirements, the need for concept re-use to support decision making becomes even more compelling.

Auto-pilot systems sense environmental changes and take appropriate action (classifiers, in the supervised machine learning context) to avoid disasters and to fly smoothly. As environmental conditions change, appropriate actions must be taken in the shortest possible time in the interest of safety. Thus for example, a situation that involves the occurrence of a sudden low pressure area coupled with high winds (a concept that would be captured by a classifier) would require appropriate action to keep the aircraft on a steady trajectory. A machine learning system that is coupled to a flight simulator can learn such concepts in the form of classifiers and store them in a repository for timely re-use when the aircraft is on live flying missions.  In live flying mode the autopilot system can quickly re-use the stored classifiers when such situations re-occur. Additionally, in live flying mode, new potentially hazardous situations not experienced  in simulator mode can also be learned and stored as classifiers in the repository for future use.

In a real world setting, there is an abundance of applications that exhibit such recurring behavior such as stock and sales applications where timely decision making results in improved productivity.  Our research setting is a data stream environment where we seek  to capture concepts as they occur, store them in highly compressed form in a repository and to re-use such concepts for classification when the need arises in the future. A number of challenges need to be overcome. Firstly, a compression scheme that captures concepts using minimal storage is required as in a high volatile high dimensional environment. Memory overhead will be a prime concern as the number of concepts will grow continuously in time given the unbounded nature of data streams. Secondly, in real-world environments, concepts rarely, if ever, occur in exactly their original form and so a mechanism is needed to recognize partial re-occurrence of concepts. Thirdly, the concept encoding scheme  needs to be efficient in order to support high speed data stream environments. 

In order to meet the above challenges, we extend the work proposed in \cite{sak:mrc} in a number of ways. In \cite{sak:mrc} concepts were initially captured using decision trees and the Discrete Fourier Transform (DFT) was applied to encode them into spectra yielding compressed versions of the original decision trees. 

Firstly, instead of encoding each concept using its own Fourier spectrum, we use an ensemble approach to aggregate individual spectra into a single unified spectrum. This has two advantages, the first of which is reduction of memory overhead.  Memory is further reduced as Fourier coefficients that are common between different spectra can be combined into a single coefficient, thus eliminating redundancy. The second advantage arises from the use of an ensemble: new concepts that manifest as a combination of previously occurring concepts already present in the ensemble have a higher likelihood of being recognized, resulting in better accuracy and stability over large segments of the data stream.

Secondly, we devise  an efficient scheme for spectral energy thresholding that directly controls the degree of compression that can be obtained in encoding concepts in the repository. 

Thirdly, we optimize the DFT encoding process by removing the need for computing a potentially expensive inner product operation on vectors.

\section{Related Research}
\label{sec:relatedresearch}
While a vast literature on concept drift detection exists \cite{pea:dci}, only a small body of work exists so far on exploitation of recurrent concepts. The methods that exist fall into two broad categories. Firstly, methods that store past concepts as models and then use a meta-learning mechanism to find the best match when a concept drift is triggered \cite{joa:trc}, \cite{gom:trc}. Secondly, methods that store past concepts as an ensemble of classifiers. The method proposed in this research belongs to the second category where ensembles remember past concepts.

An algorithm called REDDLA is presented in \cite{pli:mrc}. This algorithm is designed to handle recurring concepts with unlabeled data instances. One of the key issues is that explicit domain is required on the concept recurrence interval. The other issue is high memory overhead.

Lazarescu in \cite{laz:aml} proposed an evidence forgetting mechanism based on a multiple window approach and a prediction module to adapt classifiers based on estimating future rate of change. Whenever the  difference between observed and estimated rates of change are above a threshold, a classifier that best represents the current concept is stored in a repository. Experimentation on the STAGGER data set showed that the proposed approach outperformed the FLORA method on classification accuracy with re-emergence of previous concepts in the stream. 

Ramamurthy and Bhatnagar \cite{ram:trc} use an ensemble approach based on a set of classifiers in a global set G. An ensemble of classifiers is built dynamically from a collection of classifiers in G, if none of the existing individual classifiers are able to meet a minimum accuracy threshold based on a user defined acceptance factor. Whenever the ensemble accuracy falls below the accuracy threshold, G is updated with a new classifier trained on the current chunk of data.

Another ensemble based approach by Katakis et al. is proposed in \cite{kat:aeo}. A mapping function is applied on data stream instances to form conceptual vectors which are then grouped together into a set of clusters. A classifier is incrementally built on each cluster and an ensemble is formed based on the set of classifiers. Experimentation on the Usenet data set showed that the ensemble approach produced better accuracy than a simple incremental version of the Naive Bayes classifier. 

Gomes et al. \cite{gom:trc}  used a two layer approach with the first layer consisting of a set of classifiers trained on the current concept, while the second layer contains classifiers created from past concepts. A concept drift detector flags when a warning state is triggered and incoming data instances are buffered to prepare a new classifier. If the number of instances in the warning window is below a threshold, the classifier in layer 1 is used instead of re-using classifiers in layer 2. One major issue with this method is validity of the assumption that explicit contextual information is available in the data stream.

Gama and Kosina also proposed a two layered system in \cite{joa:trc} which is designed for delayed labelling, similar in some respects to the Gomes et al. \cite{gom:trc} approach. In their approach, Gama and Kosina pair a base classifier in the first layer with a referee in the second layer. Referees learn regions of feature space which its corresponding base classifier predicts accurately and is thus able to express a level of confidence on its base classifier with respect to a newly generated concept. The base classifier which receives the highest confidence score is selected, provided that it is above a user defined hit ratio parameter; if not, a new classifier is learnt.

Just-in-Time classifiers is the solution proposed by Allipi et al. \cite{ali:jit} to deal with recurrent concepts. Concept change detection is carried out on the classification accuracy as well as by observing the distribution of input instances. The drawback is that this model is designed for abrupt drifts and is weak at handling gradual changes.

Recently, Sakthithasan and Pears in \cite{sak:mrc} used the Discrete Fourier Transform (DFT) to encode decision trees into a highly compressed form for future use. They showed  that DFT encoding is very effective in improving classification accuracy, memory usage and processing time in general. It maintains a pool of Fourier spectra and a decision tree forest in parallel. The decision tree forest dominates the model, when none of the existing Fourier spectra matches the current concept, otherwise classification is done by the best performing Fourier spectrum.

\section{Application of the Discrete Fourier Transform on Decision Trees}
\label{sec:dftapplication}
The Discrete Fourier Transform (DFT) has a vast area of application in diverse domains such as time series analysis, signal processing, image processing and so on. It turns out as Park \cite{par:kdf} and Kargupta \cite{hhil:afs} show, that the DFT is very effective in terms of classification when applied on a decision tree model.

Kargupta et al. \cite{hhil:afs}, working in the domain of distributed data mining, showed that the Fourier spectrum fully captures a decision tree in algebraic form, meaning that the Fourier representation preserves the same classification power as the original decision tree.

\subsection{Transforming Decision Tree into Fourier Spectrum}
A decision tree can be represented in compact algebraic form by applying the DFT to paths of the tree. Each Fourier coefficient $\omega_{j}$ is given by:
\begin{equation}\\
\small
\label{coeffciientequation}
\omega_j=\frac{1}{2^d}{\sum_x{f(x)}\psi^{\overline\lambda}_j(x)};
\end{equation}
$\psi^{\overline\lambda}_j(x)=\prod_m{\exp^{\frac{2{\pi}i}{\lambda_m}{x_m}{j_m}}}$ where j and x are strings of length $d$; $x_m$ and $j_m$ represent the $m^{th}$ attribute value in $j$ and $x$respectively; $f(x)$ is the classification outcome of path vector x and $\psi^{\overline\lambda}_j(x)$ is the Fourier basis function.

\begin{figure}[h!]
\centering
\includegraphics[width=0.5\textwidth]{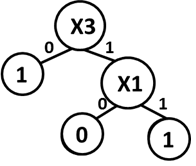}
\caption{Decision Tree with 3 binary features}
\label{fig:tree}
\end{figure}

Figure 1 shows a simple example with 3 binary valued features $x_1$, $x_2$ and $x_3$, out of which only $x_1$ and  $x_3$ are actually used in the classification.

With the wild card operator * in place we can use equations (1) and (2) to calculate non zero coefficients. Thus for example we can compute: 
\begin{align*}
\omega_{000}&=\frac{4}{8}f(**0)\psi_{000}(**0)+\frac{2}{8}f(0*1)\psi_{000}(0*1)\\
&+\frac{2}{8}f(1*1)\psi_{000}f(1*1)=\frac{3}{4}\\
\omega_{001}&=\frac{4}{8}f(**0)\psi_{001}(**0)+\frac{2}{8}f(0*1)\psi_{001}(0*1)\\
&+\frac{2}{8}f(1*1)\psi_{001}f(1*1)=\frac{1}{4}
\end{align*} and so on.

Kargupta et al in \cite{hhil:afs} showed that the Fourier spectrum of a given decision tree can be approximated by computing only a small number of {\it low order coefficients}, thus reducing storage overhead. With a suitable thresholding scheme in place, the Fourier spectrum consisting of the set of low order coefficients is thus an ideal mechanism for capturing past concepts. 

Furthermore, classification of unlabeled data instances can be done directly in the Fourier domain as it is well known that the inverse of the DFT defined in expression \ref{coeffciientequation} can be used to recover the classification value, thus avoiding the need for expensive reconstruction of a decision tree from its Fourier spectrum. The inverse Fourier Transform is given by
\begin{equation}
\label{inversefouriertransform}
\small
f(x)=\sum_j{\omega_j\overline\psi^{\overline\lambda}_j(x)} 
\end{equation}
where $\overline\psi^{\overline\lambda}_j(x)$ is the complex conjugate of $\psi^{\overline\lambda}_j(x)$.
`
An instance can be transformed into binary vector through the symbolic mapping between the actual attribute value and mapped value ( either 0 or 1 in binary case). It can then be classified using the inverse function in equation \ref{inversefouriertransform}. Suppose the instance is 010, the classification value $f(010)$ can be calculated as follows:
\small

\begin{align}
f(010)=&\frac{1}{2^d}(-1)^{000.010} \omega_{000} + \frac{1}{2^d}(-1)^{001.010}\omega_{001} \nonumber \\
&+ \frac{1}{2^d}(-1)^{010.010}\omega_{010} + \frac{1}{2^d}(-1)^{011.010}\omega_{011} \nonumber \\
&+ \frac{1}{2^d}(-1)^{100.010}\omega_{100}+ \frac{1}{2^d}(-1)^{101.010}\omega_{101} \nonumber \\
&+\frac{1}{2^d}(-1)^{110.010}\omega_{110} + \frac{1}{2^d}(-1)^{111.010} \omega_{111} = 1
\end{align}
\normalsize

\section{Exploitation of the Fourier Transform for Recurrent Concept Capture}
\label{sec:exploitationofdft}
We first present the basic algorithm used in Section \ref{subsec:ep} and then go on to discuss an optimization that we used for energy thresholding in Section \ref{subsec:dftoptimization}.

We use CBDT \cite{hoe:acb} as the base classifier which maintains a forest of Hoeffding   Trees \cite{ped:mhs} CBDT is dynamic in the sense that it can adapt to changing concepts at drift detection points.
\begin{figure}[h]
\includegraphics[width=\textwidth]{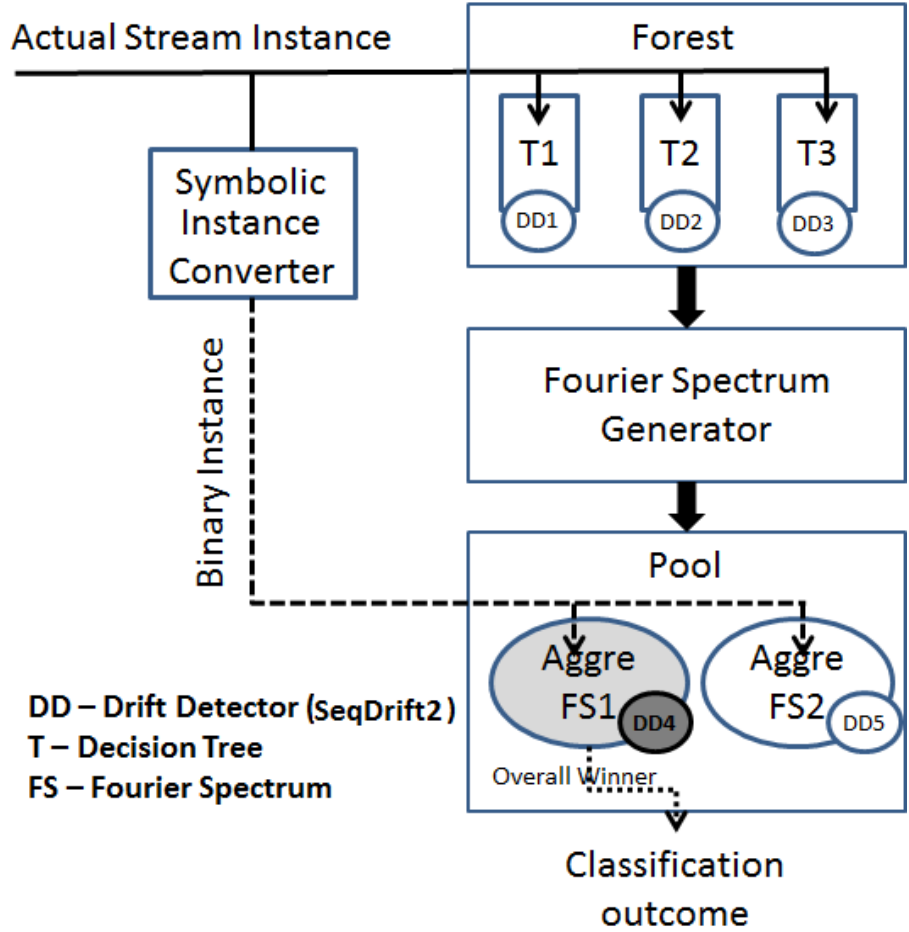}
\caption {An Architecture for Recurrent Concept Capture}
\label{fig:eparchitecture}
\end{figure}

As shown in Figure \ref{fig:eparchitecture}, the memory is divided into two segments: the forest of Hoeffding trees; and a pool of Fourier Spectra. The forest learns and undergoes structural modification on a continuous basis. The pool maintains a collection of Fourier Spectra encoded from Hoeffding Trees, each of which had the best classification accuracy across the forest at a particular concept drift point. Each Hoeffding Tree and Fourier Spectrum is equipped with an instance of a drift detector. In this research, we use the SeqDrift2 drift detector \cite{pea:dci} as the default option.

In \cite{sak:mrc}, each Fourier spectrum is represented individually as a Fourier Concept Tree (FCT). In this work, we aggregate spectra  and maintain a pool of ensemble spectra known as Ensemble Pool (EP). The aggregation process is carried out in two different ways. Algorithm $EP_a$ aggregates with reference to similarity based on accuracy whereas $EP$ aggregates based on structural similarity.  We describe the $EP$ process in Algorithm {\em EP} and discuss how FCT can be generated from it as a special case. 

In practice, any incremental decision tree approach that uses a forest of  trees can be used in place of CBDT base classifier.
\subsection{EP Algorithm}
\label{subsec:ep}
\begin {figure} [h]
\small
\begin{algorithm}{EP}{
\qinput{Energy Threshold {\it $E_T$} , Accuracy Tie Threshold ${\tau}$}
\qinput {Structural Similarity Threshold  ${\alpha}$}
\qoutput{Best Performing Classifier \textbf{C} that suits current concept}}
Plant a Hoeffding tree rooted on each attribute found in the data stream \\
\textbf{C} is set to a randomly selected Hoeffding tree model from forest \\
Initialise an empty pool \\
Read an instance \textit{I} from the data stream \\
\qrepeat \\
	Apply all classifiers in forest and pool to classify \textit{I} \\
	Append \textit{0} to the embedded drift detector's window for each classifier if classification is correct, else \textit{1}
\quntil {Drift is detected by the current best classifier \textbf{C}}\\
\qif {\textbf{C} is from the forest}\\
	Identify best performing Fourier Spectrum \textbf{F} in pool \\
	\qif{(accuracy(\textbf{C})-accuracy(\textbf{F}))$>\tau$} \\
		Apply DFT on model \textbf{C} to produce Fourier Spectrum \textbf{F*} using energy threshold {\it $E_T$} \\
		\qif{\textbf{F*} is not already in pool}\\
			Call {\it Aggregation} \qfi \qfi \qfi \\
Identify best classifier \textbf{C} across forest and pool \\
GoTo  4
\label{alg:ep}	
\end{algorithm}

\begin{algorithm}{Aggregation}{
\qinput{Fourier Spectrum \textbf{F*},a set of existing ensembles \textbf{E} in pool}
\qoutput{Updated Pool}}
\qrepeat { Over all data instances \textit{i}}\\
\qfor {Each ensemble \textbf{E} in the pool } \\
		$d(E)=d(E)+ |c(\textbf{F*},i)-c(E,i)|$ 
\qend 
\quntil {Next concept drift point} \\
$E*=\underset{E}{\operatorname{arg\,min}}(distance(E))$ \\
\qif {(E* $ \ge \alpha$)  \text{Merge \textbf{F*} with E}} \\ \qelse {insert E as a new spectrum}
\label{alg:aggregation}
\end{algorithm}
\end{figure}
\normalsize

In step 1, a Hoeffding Tree rooted on each attribute is created. In step 2, a  tree is randomly chosen as the best performing classifier \textbf{C}.  Next, an empty pool is created in step 3.  Each incoming instance is routed to all trees in the forest and pool until a concept drift signal is triggered by the drift detector instance  attached to the best classifier \textbf{C} (steps 4 to 8). At the first concept drift point,  the best performing tree \textbf{C} (in terms of drift detector's estimate of accuracy) is transformed into a Fourier Spectrum \textbf{F*} after energy thresholding \cite{sak:mrc}. In this method, the assumption is that the best tree that has the highest accuracy helps locate conceot changes precisely than other trees because it is the tree that captures concepts at a greater detail than others, thus the highest accuracy. Thereafter, \textbf{F*} is stored in the repository for reuse whenever the concept recurs. The spectra stored in the repository are  fixed in nature as the intention is to capture past concepts. A new best performing classifier is then identified as shown in step 15.

At each subsequent drift point, if the best classifier is from the pool then that classifier is applied to classify data instances until a new best classifier emerges at a subsequent drift point.  Otherwise, if the best classifier is from forest, two tests are  made prior to applying the DFT to reduce redundancy in the pool.  Firstly, we check whether the difference in accuracy between the best  Hoeffding tree in forest (C) and the best performing Fourier Spectrum (F) in the pool (from step 10) is greater than a user defined tie threshold $\tau$ (step 11). If this test succeeds, the DFT is applied to C to produce (F*) (step 12). Furthermore, a second test is made to ensure that its Fourier representation (F*) is not already in the pool (step 13). If this test is also passed, algorithm {\it Aggregation} is called to integrate F* into a selected existing Fourier Spectrum (E*) or plant (F*) as a separate Fourier spectrum in the pool (step 14).

Algorithm {\it Aggregation}  searches for the spectrum (E*) that has the greatest structural similarity to the currently generated spectrum (F*) (step 3).  Step 3 evaluates the degree of disagreement (d) between the classification decisions (c) for F* and E on data instance i. Degree of disagreement between (F*) and each of the existing ensemble (E) in pool can easily be updated incrementally in Algorithm {\it EP} using a single counter variable at each ensemble E. This removes the steps from 2 to 4 in Algorithm {\it Aggregation}.   As an alternative to aggregating {\it structurally similar  spectra}, we used accuracy as the measure that defines similarity. Similarity based on accuracy leads to aggregating similar performing Fourier Spectra together. Thus, we test the hypothesis, \textit{aggregation of two spectra based on structural similarity produces better performing trees than the one based on accuracy}. 

As stated earlier, FCT  omits the call to Algorithm {\it Aggregation} and inserts (F*) as it is, and is thus a special case of EP. 

\subsection{Optimising the Energy Thresholding Process}
\label{subsec:dftoptimization}
Sakthihasan et al. in \cite{sak:mrc} showed that classification accuracy is sensitive to spectral energy, which is given by the total of the sum of squares of the coefficients\cite{hhil:afs}); the higher the energy the greater is the classification accuracy in general. Thresholding on  spectral energy is thus an effective method of obtaining a compact spectrum while retaining the classification power inherent in the decision tree counterpart.
 
A solution described in  \cite{sak:mrc} was to iterate through each order of the spectrum and compute 
ratio of energy at orders $i-1$ to that of  $i$ respectively. Thresholding can then be implemented at order O when the ratio is less than some small tolerance value, say $0.01$. The drawback of this simple solution is that it does not guarantee that the cumulative energy up to order O contains a proportion ($\epsilon$) of the total energy. Fortunately, a solution exists for this problem. Theorem 1 proves that E(T) (total energy of Fourier Spectrum) equals to $\omega_{\overline 0}$ (The $0^{th}$ coefficient). Thus, total energy can be computed efficiently, without having to enumerate all the single coefficients. \\ \\

\textbf{Theorem 1}
The total spectral energy $E =\sum_j{\omega_{{j}}^2} = \omega_{\overline 0}$, where  $\omega_{\overline 0}$ denotes the coefficient with order 0, which is easily computed as its Fourier basis function is unity. \\ \\
\textbf{Proof:} For case 1 we prove the result when exactly one such combination exists and discuss the extension to the case when more than one combination is present. Without loss of generality we illustrate the proof when the wild card characters occur at the beginning of vector x; if they occur in any other position then a simple reordering operation can be used without affecting the validity of the proof. \\ \\
Suppose that the cardinality of the attributes after reordering are $\lambda_i$ where $i \in [0,d-1]$, where d is the dimensionality of the dataset.
\small
\begin{align}
 \sum_{x \in S}\psi_j(x)&=\exp \Bigl(\frac{2{\pi}i{j_0}0}{\lambda_0} \Bigr) \times \exp \Bigl(\frac{2{\pi}i{j_1}{x_1}}{\lambda_1} \Bigr) \times \cdots \times \exp \Bigl(\frac{2{\pi}i{j_{d-1}}x_{d-1}} {\lambda_{d-1}}\Bigr) \nonumber \\
&+\exp\Bigl(\frac{2{\pi}i{j_0}1}{\lambda_0}\Bigr) \times \exp\Bigl(\frac{2{\pi}i{j_1}{x_1}}{\lambda_1}\Bigr) \times \cdots \times \exp\Bigl(\frac{2{\pi}i{j_{d-1}}{x_{d-1}}}{\lambda_{d-1}}\Bigr) \nonumber \\
&\vdots \nonumber \\
&+\exp\Bigl(\frac{2{\pi}i{j_0}(\lambda_0-1)}{\lambda_0}\Bigr) \times \exp\Bigl(\frac{2{\pi}i{j_1}{x_1}}{\lambda_1}\Bigr) \times \cdots \times \exp\Bigl(\frac{2{\pi}i{j_{d-1}}{x_{d-1}}}{\lambda_{d-1}}\Bigr) \nonumber \\
&= \exp \Bigl(\frac{2{\pi}i{j_1}{x_1}}{\lambda_1} \Bigr) \times \cdots \times \exp \Bigl(\frac{2{\pi}i{j_{d-1}}x_{d-1}} {\lambda_{d-1}}\Bigr)\sum_{k=0}^{\lambda_0-1}\Bigl(\exp \Bigl(\frac{2{\pi}i{j_0}k}{\lambda_0} \Bigr)\Bigr) \nonumber \\
&= \exp \Bigl(\frac{2{\pi}i{j_1}{x_1}}{\lambda_1} \Bigr) \times \cdots \times \exp \Bigl(\frac{2{\pi}i{j_{d-1}}x_{d-1}} {\lambda_{d-1}}\Bigr)\sum_{k=0}^{\lambda_0-1}\Bigl(\psi_o\psi_k\Bigr) \mbox { as $\psi_0=1$} \nonumber \\
&=0 \mbox{ due to orthogonality of the Fourier basis functions} \nonumber
\end{align}
\normalsize
When n wild cards are present, the extension is straightforward and results in:
\small
\begin{align}
\sum_{x \in S}\psi_j(x)&= \exp \Bigl(\frac{2{\pi}i{j_{n}}{x_{n}}}{\lambda_n} \Bigr) \times \cdots \times \exp \Bigl(\frac{2{\pi}i{j_{d-1}}x_{d-1}} {\lambda_{d-1}}\Bigr)  \prod_{l=0}^{n-1}\sum_{k=0}^{\lambda_l-1}\Bigl(\exp \Bigl(\frac{2{\pi}i{j_l}k}{\lambda_l} \Bigr)\Bigr)  \\ 
 \sum_{x \in S}\psi_j(x)&= \exp \Bigl(\frac{2{\pi}i{j_{n}}{x_{n}}}{\lambda_n} \Bigr) \times \cdots \times \exp \Bigl(\frac{2{\pi}i{j_{d-1}}x_{d-1}} {\lambda_{d-1}}\Bigr)  \prod_{l=0}^{n-1}\sum_{k=0}^{\lambda_l-1}\Bigl(\psi_0\psi_k\Bigr) \nonumber \\ 
&=0 \mbox{ due to orthogonality of Fourier basis functions } \nonumber
\end{align}
\normalsize
This optimization significantly increases processing speed, especially in high dimensional data stream environments. \\ \\
The next optimization is applied to optimize the Fourier Basis function calculation in equation \ref{coeffciientequation} especially when wildcard characters (denoting absence of a feature) are present in a path vector $x$ of a Hoeffding Tree. 

\subsection{Optimizing the Computation of the Fourier Basis Function}
The computation of a Fourier basis function for a given partition  $j$ in a generic $n-ary$ ( $n\ge2$) domain is given by:
\begin{align}
 \sum_{x \in S}\psi_j(x)&=\sum_{x \in S}\prod_m{\exp^{\Bigl(\frac{2{\pi}i{j_m}{x_m}}{{\lambda_m}}\Bigr)}} 
\end{align}
Thus we can see from (4.3) that the computation of $ \sum_{x \in S}\psi(j)$ over a set of schema S requires the computation of an  expensive inner product operation between the $x$ and $j$. However, it is possible to optimize this inner product computation as defined in Theorem 2. \\ \\ \textbf{Theorem 2} 
The computation of $ \sum_{x \in S}\psi_j(x)$ can be optimized as follows: \\ \\
Case 1: If there exists at least one $(p,*)$ combination with  $p \in j$, $p\ne 0$  and $*$  a wild card character defining a set of schema $S$, then $ \sum_{x \in S}\psi_j(x)=0$. \\ \\
Case 2: else if there exists $n$ combinations of $(0,*)$ pairs in the $j$ and $x$ vectors respectively, then  
\[
 \sum_{x \in S}\psi_j(x)= {\lambda}\prod_{k=n}^{\lambda_k-1}{\exp^{\Bigl(\frac{2{\pi}i{j_k}{x_k}}{{\lambda_k}}\Bigr)}}  \mbox{ where $\lambda=\prod_{l=0}^{n-1}{\lambda_l}$}
\]
\textbf{Proof:} For case 1 we prove the result when exactly one such combination exists and discuss the extension to the case when more than one combination is present. Without loss of generality we illustrate the proof when the wild card characters occur at the beginning of vector x; if they occur in any other position then a simple reordering operation can be used without affecting the validity of the proof. \\ \\
Suppose that the cardinality of the attributes after reordering are $\lambda_i$ where $i \in [0,d-1]$, where d is the dimensionality of the dataset.
\small
\begin{align}
 \sum_{x \in S}\psi_j(x)&=\exp \Bigl(\frac{2{\pi}i{j_0}0}{\lambda_0} \Bigr) \times \exp \Bigl(\frac{2{\pi}i{j_1}{x_1}}{\lambda_1} \Bigr) \times \cdots \times \exp \Bigl(\frac{2{\pi}i{j_{d-1}}x_{d-1}} {\lambda_{d-1}}\Bigr) \nonumber \\
&+\exp\Bigl(\frac{2{\pi}i{j_0}1}{\lambda_0}\Bigr) \times \exp\Bigl(\frac{2{\pi}i{j_1}{x_1}}{\lambda_1}\Bigr) \times \cdots \times \exp\Bigl(\frac{2{\pi}i{j_{d-1}}{x_{d-1}}}{\lambda_{d-1}}\Bigr) \nonumber \\
&\vdots \nonumber \\
&+\exp\Bigl(\frac{2{\pi}i{j_0}(\lambda_0-1)}{\lambda_0}\Bigr) \times \exp\Bigl(\frac{2{\pi}i{j_1}{x_1}}{\lambda_1}\Bigr) \times \cdots \times \exp\Bigl(\frac{2{\pi}i{j_{d-1}}{x_{d-1}}}{\lambda_{d-1}}\Bigr) \nonumber \\
&= \exp \Bigl(\frac{2{\pi}i{j_1}{x_1}}{\lambda_1} \Bigr) \times \cdots \times \exp \Bigl(\frac{2{\pi}i{j_{d-1}}x_{d-1}} {\lambda_{d-1}}\Bigr)\sum_{k=0}^{\lambda_0-1}\Bigl(\exp \Bigl(\frac{2{\pi}i{j_0}k}{\lambda_0} \Bigr)\Bigr) \nonumber \\
&= \exp \Bigl(\frac{2{\pi}i{j_1}{x_1}}{\lambda_1} \Bigr) \times \cdots \times \exp \Bigl(\frac{2{\pi}i{j_{d-1}}x_{d-1}} {\lambda_{d-1}}\Bigr)\sum_{k=0}^{\lambda_0-1}\Bigl(\psi_o\psi_k\Bigr) \mbox { as $\psi_0=1$} \nonumber \\
&=0 \mbox{ due to orthogonality of the Fourier basis functions} \nonumber
\end{align}
\normalsize
When n wild cards are present, the extension is straightforward and results in:
\small
\begin{align}
\sum_{x \in S}\psi_j(x)&= \exp \Bigl(\frac{2{\pi}i{j_{n}}{x_{n}}}{\lambda_n} \Bigr) \times \cdots \times \exp \Bigl(\frac{2{\pi}i{j_{d-1}}x_{d-1}} {\lambda_{d-1}}\Bigr)  \prod_{l=0}^{n-1}\sum_{k=0}^{\lambda_l-1}\Bigl(\exp \Bigl(\frac{2{\pi}i{j_l}k}{\lambda_l} \Bigr)\Bigr)  \\ 
 \sum_{x \in S}\psi_j(x)&= \exp \Bigl(\frac{2{\pi}i{j_{n}}{x_{n}}}{\lambda_n} \Bigr) \times \cdots \times \exp \Bigl(\frac{2{\pi}i{j_{d-1}}x_{d-1}} {\lambda_{d-1}}\Bigr)  \prod_{l=0}^{n-1}\sum_{k=0}^{\lambda_l-1}\Bigl(\psi_0\psi_k\Bigr) \nonumber \\ 
&=0 \mbox{ due to orthogonality of Fourier basis functions } \nonumber
\end{align}
\normalsize
The value of Case 1 is that a simple scan of the $j$ and $x$ vectors will save a total of $d$ multiplications and $d-1$ additions. \\ \\
We now turn our attention to Case 2. Since $\prod_l{\lambda_l}$ is a constant for all possible values of $j$ and $y$, the value of Case 2 is that a scan of the two vectors will avoid the overhead of $n$ multiplications and $n-1$ additions. 

Even with these optimizations, coefficient calculation may be expensive in a large dimensional data set. In the next section we present a strategy to further optimize the derivation of the spectrum.

\subsection{Localized Approach to Ensemble Learning in the Fourier Domain}
\label{sec:agg}
In order to realize the full benefits of ensemble learning in the Fourier domain,  we aggregate individual spectra $s_i(x)$ that represent different concepts which manifest at different points in the stream. 
\begin{align}
\label{eqn:agg1}
s_c(x)&=\sum_i{A_i\sum_i{s_i(x)}} \nonumber \\
          &=\sum_i{A_i\sum_{j \in P_i}{\omega_j}^{(i)}\overline\psi_{{{j}}(x)}} 
\end{align}
where $s_c(x)$ denotes the ensemble spectrum produced from the individual spectra $s_i(x)$ produced at different points $i$ in the stream; $A_i$ is the classification accuracy of its corresponding spectrum and $P_i$ is the set of partitions for non zero coefficients in spectrum $s_i$.

Park in \cite{par:kdf} used ensemble learning with Fourier spectra in a setting different to ours. They considered a distributed system with each node $i$ producing its own spectrum $s_i(x)$ and aggregation taking place at a central node. In our setting of a data stream environment, we do not have all spectra in advance but we can still use the same principle due to the distributive nature of the linear weighted sum expressed by $(\ref{eqn:agg1})$. Hence, we use:
\begin{equation}
\label{eqn:agg2}
s_c^{(i+1)}(x)=s_c^{(i)}(x)+A_{i+1}{s_{i+1}(x)}
\end{equation}
where  $s_c^{(i+1)}(x)$, $s_c^{(i)}$ represent the ensemble spectra at concept drift points $i+1$ and $i$ respectively in the stream and $s_{i+1}(x)$ is the spectrum produced at drift point $i+1$ with accuracy $A_{i+1}$. 

We use expression $(\ref{eqn:agg2})$ for implementing ensemble learning but with one essential difference. A direct application of $(\ref{eqn:agg2})$ using the entire (global) set of attributes $G$ comprising the data set would be inefficient. As there are an exponential number of coefficients with respect to the number of attributes, this could cause a bottleneck in high dimensional environments. One practical solution is to populate the spectrum using only attributes present in a given tree. The  major advantage of this approach is smaller computational overhead as the Fourier transform effort is directly proportional to the size of the attribute set used. Then this initial spectrum can be extended to a full length spectrum containing the attributes that are absent in the given tree, using a  simple transformation scheme.

We define an attribute set of a Decision Tree as that subset of attributes which define splits in the tree. Suppose that we are integrating spectra from  trees $D_1$ and $D_2$, having attribute sets $L$ and $M$ respectively. We apply the DFT on $D_1$ to obtain $S_1$ using \emph{only the attributes in its attribute set $L$ and not all attributes in $G$}.  Similarly we generate $S_2$ from $D_2$ using only the attributes defined in $M$. 

Now, in order to integrate $S_1$ with $S_2$, we need to account for differences in the attribute sets $L$ and $M$. To do this, we take $S_1$ and expand the spectrum by incorporating attributes in the set $M \setminus L$. The expansion is defined by a single operation:

For each schema instance in the spectrum (say $S_1$) expand the spectrum by adding $0$ to all attribute index positions  in set $M \setminus L$. The {\it coefficient value after expansion will remain it the same as the classification $f$} value for all of these added index positions remains unchanged.
We are now in a position to integrate two spectra produced from their own localized set of attributes. Essentially, this means that we now have a more efficient method of implementing ensemble learning using expression $(\ref{eqn:agg2})$.

The next section presents the empirical outcomes of the proposed models  with the above mentioned optimizations.
\section{Experimental Study}
\label{sec:experimentalstudy}
The main focus of the study is to assess the effectiveness of the ensemble EP approach  vis-a-vis  FCT  in respect of classification accuracy, memory consumption, processing speed, tolerance to noise. We also assessed the sensitivity of EP's accuracy on two significant factors, pool size and impact of drift detector. All experimentation was done with the following parameter values: \\ \\
      \emph{Tree Forest:}
                Max Node Count=5000, Max Number of Fourier spectra=10, Tie Threshold $\tau$=0.01
      \\ \emph{SeqDrift2/ADWIN \cite{bit:lft}:} drift significance value=0.01 

\subsection{Datasets Used for the Experimental Study}
\label{sec:datasets}
\subsubsection{Synthetic Data}
We experimented with the Rotating Hyperplane data generator that is commonly used in drift detection and recurrent concept mining. The dataset was   generated within the MOA data stream tool \cite{bit:moa}. We injected concept recurrence into the stream at known points so that we could evaluate the capabilities of FCT and EP to recognize and exploit such recurrences. 
For this dataset 10 different concepts were generated, each of which spanned 5,000 instances and each occurred a total of 3 times at different points in the stream. In order to challenge the concept recognition process, we added 10\% noise by inverting the class labels of 10\% of randomly selected instances.

\subsubsection{Real World Data} 
{ \it Spam Data Set: }  The Spam dataset was used in its original form \footnote{from {\it http://www.liaad.up.pt/kdus/products/datasets-for-concept-drift}} which encapsulates  an evolution of Spam messages. There are 9,324 instances and 499 informative attributes. 

{\it Electricity Data Set: }
NSW Electricity dataset is also used in its original form \footnote{from {\it http://moa.cms.waikato.ac.nz/datasets/}}. There are two classes {\it Up} and {\it Down} that indicate the change of price with respect to the moving average of the prices in last 24 hours.

{\it Flight Data Set:}
This dataset is generated through the use of NASA's FLTz flight simulator which was designed to simulate flight conditions experienced with commercial flights. It consists of a set of 20 separate files, each containing data about a single flight with four scenarios: take off, climb, cruise and landing. Data is recorded every second and a data instance is produced.  The "Velocity" feature is chosen as the class feature as it needs to be adjusted in order to maintain aircraft stability during various maneuvers such as take off and landing. Velocity was discretized into binary outcomes "UP" or "DOWN" depending on the directional change of the moving average in a window of size 10 data instances.

\subsection{Comparative Study:  Ensemble versus Single Spectrum Approach}
Previous research on the use of Fourier spectrum revealed accuracy and memory advantages over meta learning approaches such as the one employed by Gama and Kosina in storing past concepts in a repository \cite{sak:mrc}. For details of the advantages of the Fourier approach and experimentation with it the reader is referred to  \cite{sak:mrc}. Our focus here is a comparative study of  the Ensemble approach versus the single spectrum approach. With this in mind we designed three types of experiments.

\subsubsection{Accuracy}
Accuracy is a critical performance measure in many practical applications. 
Due to the dynamic nature of data streams classification accuracy on the current concept was taken as the performance measure.
\begin{figure}[h]
\setlength{\abovecaptionskip}{6pt plus 3pt minus 2pt}
\includegraphics[width=\textwidth]{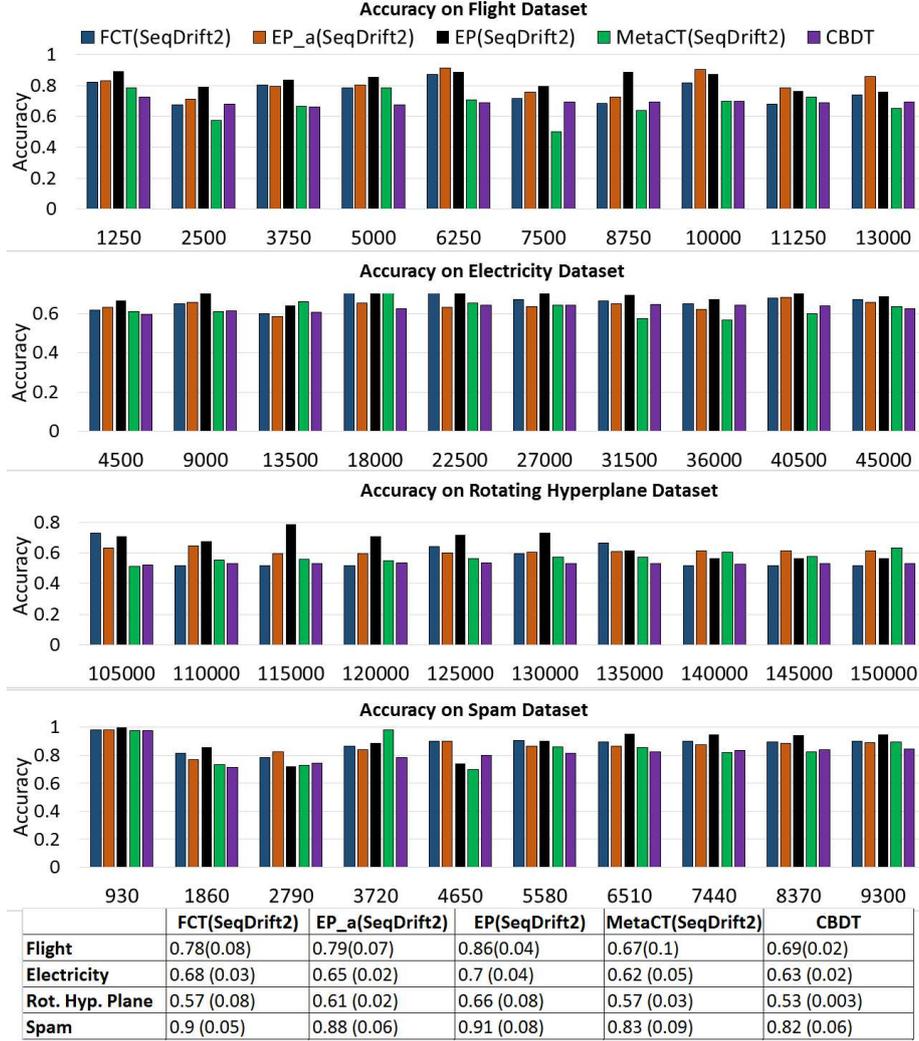}
\caption{Accuracy Profiles}
\label{accuracycurves}
\end{figure}
Figure \ref{accuracycurves} presents  Accuracy values of all algorithms at 10 equal-sized sub-divisions of the stream. 
We also present overall mean and standard deviations of accuracy taken across the entire stream for each dataset.

Fig  \ref{accuracycurves} shows that the individual accuracies across segments and overall accuracy across the entire stream are consistent  with each other. MetaCT, which uses a referee based strategy was found to be the worst performing algorithm on all datasets. In contrast EP outperforms the other algorithms in general, followed by $EP_a$ and FCT. These results show clearly that DFT based methods are superior in a dynamic data stream environment.

FCT does not exploit aggregation of Fourier Spectra and is hence challenged in a memory constrained environment where the number of models stored for reuse is limited. Figure \ref{accuracycurves} depicts the performance in such an environment where memory is severely limited. This introduces a large burden on FCT to re-learn concepts after change. 
EP is more resilient at small pool sizes as any given concept that recurs can be approximated by a linear combination of spectra embedded in the ensemble, just as a waveform of arbitrary shape can be approximated by a large enough sum of sine functions in signal processing.

Examining model usage statistics, EP was 3.7 times higher in model re-use on the Flight dataset. The corresponding value was 2.6 for $EP_a$ on the same dataset. This provides empirical support for the claim that an aggregation-based model such as EP has a significant advantage in reducing the degree of relearning. 
For Rotating Hyperplane with known recurrence points the advantage of EP over its counterparts is very explicit. We display the stream segment for the third round of concept occurrences, spanning the 10 concepts. Each of the 10 intervals represent the second recurrence of a concept and the Figure shows that EP outperforms FCT on 8/10 concepts; $EP_a$ and MetaCT on 7/10 concepts; and CBDT on all 10 concepts.
The next key aspect in a memory constrained environment is memory consumption which is assessed in the following section.

\subsubsection{Memory}
Memory consumption is influenced by the degree of generalizability of a given algorithm. A greater degree of generalizability promotes higher re-use and reduces the number of spectra that need to be stored in the repository to achieve a given level of classification accuracy. In this context it will be interesting to compare the consumption of EP with that of FCT as they have contrasting model re-use characteristics.

MetaCT(SeqDrift2) and CBDT were excluded from memory comparison due to their relatively poor performance in the previous experiment.
\begin{center}
\begin{table} [h]
\caption{Memory Usage with Pool size set to 10}
\centering
  \begin{tabular}{| l | r | r | r | r |}
    \hline 
Dataset & \multicolumn{3}{c|}{Average Pool Memory (in KBs)} \\ \hline
& FCT & $EP_a$ & EP \\ \hline
Flight & 32.1 & 20.2 & {\bf 18.1} \\ \hline
Electricity & 31.6 & 16.1 & {\bf 14.1} \\ \hline
Rot. Hyperplane & 48.4 & 38.6 & {\bf 27.9} \\ \hline
Spam & 17.3 & 17.2 & {\bf 16.4} \\ \hline
  \end{tabular}
\label{tab:memorycomparison}
\end{table}
\end{center}
Table \ref{tab:memorycomparison} presents the average memory consumption of the pool over the entirety of each dataset. 
As mentioned in Section \ref{sec:dftapplication}, each of the above algorithms in Table \ref{tab:memorycomparison} has two components: a forest and a repository pool. Memory consumed by forest is not a distinguishing factor as there was a very marginal difference between the  algorithms and thus the focus was on the repository pool.

Without exception, EP consumed the least memory compared to the other  algorithms. This was expected as EP structurally examines instance vectors (i.e. corresponding to classification paths in Hoeffding tree) and aggregates similar vectors together. On the other hand in $EP_a$, structural similarity is not guaranteed and two structurally very different  Spectra producing similar accuracy could be chosen as the candidates to be aggregated, thus resulting in larger spectra.  Table \ref{tab:memorycomparison} provides evidence to support this premise as the memory consumed by $EP_a$ is higher than that of EP but lower than FCT. On average over all  datasets, EP achieved a 41\% reduction in memory consumption in relation to FCT; the corresponding figure for Electricity was  55\%. This represents a  significant benefit of applying aggregation in Fourier space.

\subsubsection{Processing Speed}
DFT application is a potential performance bottleneck when compared to classification, especially in high dimensional data streams. 

Processing speed is dependent on a variety of factors: maintaining and classifying relatively larger number of Fourier Spectra in FCT compared to EP and $EP_a$, aggregation in EP and $EP_a$ that generalize models thus reducing re-learning and the need for DFT application, and finally the computational overheads of aggregation. Therefore, this section assesses the trade off between single and aggregated Fourier approaches in terms of processing speed. 
\begin{center}
\begin{table} [h]
\caption{Processing Speed in instances per second}
\centering
  \begin{tabular}{| l | r | r | r |}
    \hline 
Dataset & FCT& $EP_a$& EP \\ \hline
Flight & 797.2 & 731.2 & \textbf{836.9} \\ \hline 
Electricity & \textbf{11600.3} & 9002.5 & 11402.5 \\ \hline
Rotating Hyperplane & 5647.8 & 5413.8 & \textbf{5804.5} \\ \hline
Spam & \textbf{4.2} & 3.9 & \textbf{4.2} \\ \hline
  \end{tabular}

\label{tab:processingspeed}
\end{table}
\end{center}
Table \ref{tab:processingspeed} shows that EP is the fastest most of the time. 
$EP_a$, even though it has the potential to be faster due to its simple aggregation strategy, suffers from inappropriate aggregations that introduce instability, thus triggering more drift points than its EP counterpart. EP, on the other hand, efficiently does structural similarity comparison by incrementally updating simple counters that remembers the number of disagreements in classification between the current winner tree and every Fourier Spectra in pool. On the other hand, although EP through its aggregation strategy requires more computational effort than $EP_a$, that effort is compensated for by its stability, which triggers fewer false drift alarms than either $EP_a$ or FCT.  Therefore, this experiment demonstrates that an expensive operation such as aggregation if applied appropriately will yield a direct processing speed advantage over a period of time.

\subsubsection{Effects of Noise} 
Algorithms that work well in noise-free environments will fail on noisy environments if they lack the ability to generalize to new data by removing minor variations which often correspond to noise. DFT application, as mentioned earlier, extracts significant coefficients by ignoring minor coefficients that may capture noise inherent in data. It was shown in \cite{sak:mrc} that DFT application  provides robustness in a noisy environment as opposed to a non-DFT based approach such as MetaCT. Therefore, this experiment is aimed at testing whether aggregation has an added advantage over a non-aggregation based method such as FCT. 

\begin{figure}[h]
\setlength{\abovecaptionskip}{6pt plus 3pt minus 2pt}
\includegraphics[width=\textwidth]{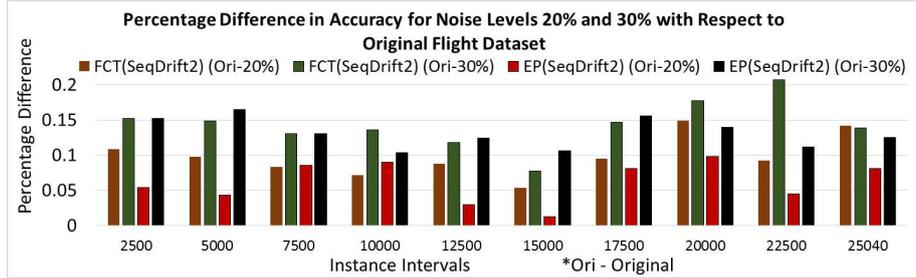}
\caption{The impact of noise on accuracy}
\label{noiseaccuracy}
\end{figure}

Figure \ref{noiseaccuracy} shows percentage  accuracy decrease for noise levels 20\% and 30\% on  FCT and EP relative to accuracy on the original Flight dataset. It is clear that the decrease in accuracy is higher at the 30\% noise level. What is interesting is the higher tolerance of EP to noise compared to FCT. In 8/10 intervals, for 20\% noise, EP is found to be having a lesser decrease than its counterpart. Similarly at the 30\% noise level, the fraction is 4/10, with the two being tied in performance in two other intervals. Again, as with the other metrics that we tracked, the superior performance of EP can be explained in terms of its power to generalize making it more robust to the effects of noise \footnote{The other 3 datasets that we experimented with displayed similar trends  to that of the Flight dataset and were thus not included in interests of space constraints}.

Next we examine the sensitivity of EP on key parameters that significantly affect performance. Due to the superiority of EP over the other algorithms, the study was confined to this algorithm. Please refer \cite{sak:mrc} for sensitivity analysis on FCT's parameters.

\subsection{Sensitivity Analysis}
EP(SeqDrift2) has two key parameters of its own: pool size and choice of drift detector. 


\subsubsection{Pool Size}
In this experiment we contrasted classification accuracy at two different ends of the pool size scale, namely 1 and 10. In the context of the Flight dataset which has four concepts, a pool size of 1 represents an extremely limiting memory environment and the size of 10 represents a situation where memory is plentiful. 
\begin{figure}[h]
\setlength{\abovecaptionskip}{6pt plus 3pt minus 2pt}
\includegraphics[width=\textwidth]{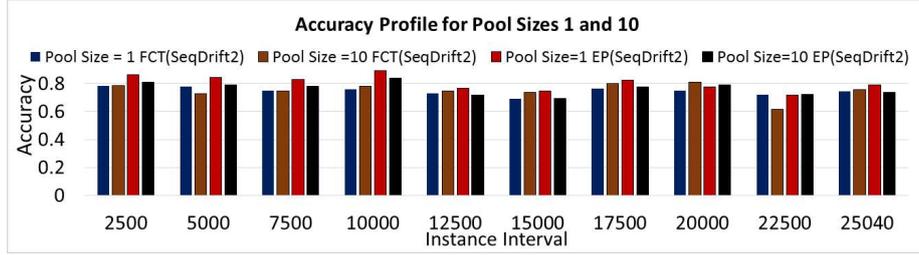}
\caption{The impact of pool size on flight dataset}
\label{poolsizeaccuracy}
\end{figure}
Figure \ref{poolsizeaccuracy} shows accuracy values over 10 intervals. Interestingly, EP, with pool size1, has the highest accuracy in 8/10 intervals. There is a 7.6\% and 7.2\% gain in accuracy compared to FCT over pool sizes 1 and 10 respectively.  This is a significant outcome of this research. Even in an extreme memory challenged environment, EP achieves its best accuracy over a setting with a much higher memory capacity. The implication is that ensemble accuracy increases with greater diversity and resonates with the research conducted by  \cite{gas:dte}. This illustrates the strength of aggregation applied in the EP algorithm. As more memory becomes available at  pool size 10, FCT's accuracy converges to that of its counterpart, as expected. At the higher memory setting FCT can accommodate more spectra in its pool that are tailored to specific concepts.

\subsubsection{Impact of Drift Detector}
A drift detector that incorrectly triggers change points leads  to partial learning of a concept and under developed classifiers being stored in the pool. This introduces fluctuations in accuracy, which in turn trigger change detections, causing even more fluctuations and so on. This is a cyclic problem. On the other hand, if a drift detector fails to detect changes, classifiers are not updated in a timely fashion, thus leading to poor performance. This situation may arise if a drift detector   has significantly high detection delay in signaling changes. The ADWIN and SeqDrift2 drift detectors, as shown in \cite{pea:dci} have contrasting properties. SeqDrift2 has a lower false positive rate than ADWIN while having similar sensitivity to ADWIN. Therefore, the comparative study is largely governed by false positive detections. 

\begin{figure}[h]
\setlength{\abovecaptionskip}{6pt plus 3pt minus 2pt}
\includegraphics[width=\textwidth]{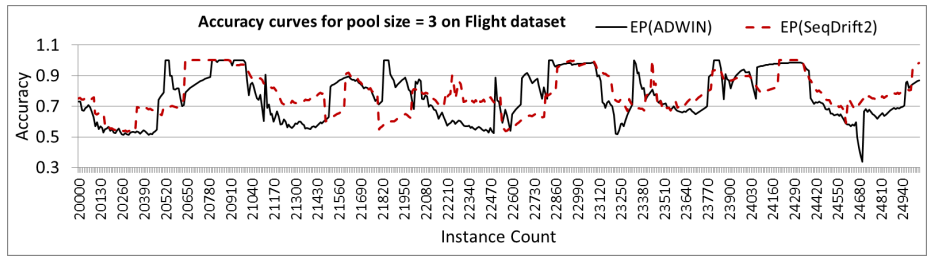}
\caption{The impact of drift detector on EP with pool size 10}
\label{driftdetectoraccuracy}
\end{figure}

Figure \ref{driftdetectoraccuracy} reveals that SeqDrift2 helped EP to reduce the frequency of sudden accuracy drops seen with ADWIN, due to the latter signaling false changes in concepts. In the segment  shown in Figure \ref{driftdetectoraccuracy}, there is a 5\% gain in accuracy by using SeqDrift2 and it is 3.4\% over the entire data set. 

\section{Conclusions and Future Work}
\label{sec:conclusion}
In this research we proposed a novel approach for capturing and exploiting recurring concepts in data streams. We  optimized the derivation of the Fourier spectrum by employing two mechanisms: one for energy thresholding and the other for speeding up  computation of the Fourier basis functions.

This research revealed that the ensemble approach outperformed the single spectrum approach and is thus the method of choice in high speed dynamic environments that generate large amounts of concepts over the progression of the stream. In such environments FCT would be challenged in terms of memory capacity and would be forced to flush portions of its repository sooner that EP, thus losing its ability to exploit concept recurrences and in turn leading to a loss of accuracy. However, as shown in the experimentation care needs to be taken on how spectra are combined: a naive approach of simply combining similarly performing spectra in terms of accuracy can be worse than maintaining single spectra. We showed that the structural similarity scheme outperformed the other two approaches on a broad set of criteria including accuracy, robustness to noise and over-fitting, memory consumption and processing speed. 

In terms of  future work there are two promising directions. We believe that is possible to further reduce the computational effort involved in deriving the spectrum by only keeping the lowest order coefficient at each leaf node of the Decision Tree together with a residual coefficient that  captures the contribution of other coefficients at that node. Secondly, at each concept drift point we can parallelize computation of the spectrum in one thread while processing incoming instances in another thread in a parallel environment such as a Spark framework.


\begin{thebibliography}{1}
\setlength{\itemsep}{1mm}

\bibitem{ali:jit}
C.~Alippi, G.~Boracchi and M.~Roveri.
\newblock {\em Just-In-Time Classifiers for Recurrent Concepts}.
\newblock IEEE Transactions on Neural Networks and Learning Systems,
  vol.~24(4), pages 620--634, 2013.

\bibitem{bit:lft}
Bifet, A. \& Gavald{\`a}, R.
\newblock {\em Learning from Time-Changing Data with Adaptive Windowing}.
\newblock In Proceedings of the 7th SIAM ICDM, pages 443--448. SIAM, 2007.

\bibitem{bit:moa}
Bifet, A. Holmes, G. Kirkby, R. \& Pfahringer, B.
\newblock {\em {MOA}: Massive Online Analysis}.
\newblock The Journal of Machine Learning Research, vol(11), pages 1601--1604, 2010.
  
\bibitem{ped:mhs}
Domingos, P. \& Hulten, G.
\newblock {\em Mining High-speed Data Streams}.
\newblock In Proceedings of the ACM SIGKDD'00,  pages 71--80, New York, NY,  USA, 2000. ACM.

\bibitem{joa:trc}
Gama, J. \& Kosina, P.
\newblock {\em Learning about the Learning Process}.
\newblock In Advances in Intelligent Data Analysis X, vol(7014) of {\em Lecture Notes in Computer Science}, pages 162--172. Springer Berlin Heidelberg, 2011.

\bibitem{gas:dte}
Gashler, M., Giraud-Carrier C., \& Martinez, T.
\newblock {\em Decision Tree Ensemble: Small Heterogeneous Is Better Than Large Homogeneous}.
\newblock {7th International Conference on Machine Learning and Applications}, pages 900--905,IEEE Computer Society, 2008.

\bibitem{gom:trc}
Gomes, J. Menasalvas, E. \& Sousa, P.
\newblock {\em Tracking Recurrent Concepts Using Context}.
\newblock In Rough Sets and Current Trends in Computing, vol(6086), pages
  168--177. Springer Berlin Heidelberg, 2010.

\bibitem{hoe:acb}
Hoeglinger, S. Pears, R. \& Koh, Y.
\newblock {\em {CBDT}: A Concept Based Approach to Data Stream Mining}.
\newblock In Proceedings of the PAKDD '09, pages 1006--1012, Berlin,
  Heidelberg, 2009. Springer-Verlag.



\bibitem{kat:aeo}
Katakis, I. Tsoumakas, G. \& Vlahavas, I.
\newblock {\em An Ensemble of Classifiers for Coping with Recurring Contexts in
  Data Streams}.
\newblock In Proceedings of the ECAI'08 , pages 763--764, Amsterdam, Netherlands, The Netherlands, 2008. The IOS Press.

\bibitem{hhil:afs}
Kargupta, H. Park, B. \& Dutta, H.
\newblock {\em Orthogonal Decision Trees}.
\newblock IEEE Transactions on Knowledge and Data Engineering, vol(18), no(8),
  pages 1028--1042, 2006.

\bibitem{laz:aml}
Lazarescu, M.
\newblock {\em A Multi-Resolution Learning Approach to Tracking Concept Drift
  and Recurrent Concepts}.
\newblock In 5th international workshop on Pattern Recognition in Information
  Systems, 2005.

\bibitem{par:kdf}
Park, B.
\newblock {\em Knowledge Discovery from Heterogeneous Data Streams Using
  Fourier Spectrum of Decision Trees}.
\newblock PhD thesis, Pullman, WA, USA, 2001.

\bibitem{pea:dci}
Pears, R. Sripirakas, S. \& Koh, Y.
\newblock {\em Detecting concept change in dynamic data streams}.
\newblock Machine Learning, 97:3, pp 259--293, 2014.

\bibitem{pli:mrc}
Peipei Li, Xindong Wu, and Xuegang Hu, "Mining recurring concept drifts with
    limited labeled streaming data," ACM Trans. Intell. Syst. Technol.,vol. 3, no. 2, pp. 29:1-29:32, Feb. 2012

\bibitem{ram:trc}
Ramamurthy, S. \& Bhatnagar, R.
\newblock {\em Tracking recurrent concept drift in streaming data using  ensemble classifiers}.
\newblock In 6th International Conference on Machine Learning Applications,
  pages 404--409, Dec 2007.

\bibitem{sak:mrc}
Sripirakas, S. \& Pears, R.
\newblock {\em Mining Recurrent Concepts in Data Streams Using the Discrete
  Fourier Transform}.
\newblock In DaWaK'14, vol(8646) of {\em Lecture Notes in
Computer Science}, pp 439--451. Springer International Publishing, 2014.
\end{thebibliography}

\end{document}